\documentclass[a4paper,fleqn]{cas-sc}

\usepackage[authoryear,longnamesfirst]{natbib}
\usepackage{booktabs}
\usepackage{graphicx}   
\usepackage{amsmath}

\def\tsc#1{\csdef{#1}{\textsc{\lowercase{#1}}\xspace}}
\tsc{WGM}
\tsc{QE}
\tsc{EP}
\tsc{PMS}
\tsc{BEC}
\tsc{DE}

\begin{document}
\let\WriteBookmarks\relax
\def\floatpagepagefraction{1}
\def\textpagefraction{.001}
\shorttitle{NuHF-Claw for Cognitive-Risk Procedure Support}
\shortauthors{Xiao et~al.}

\title{NuHF-Claw: A Risk-Constrained Cognitive Agent Framework for Human-Centered Procedure Support in Digital Nuclear Control Rooms}     

\tnotetext[1]{The research was supported by a grant from the National Natural Science Foundation of China (Grant No. T2192933), the Foundation of National Key Laboratory of Human Factors Engineering (Grant No. HFNKL2024W07), and Tsinghua University Initiative Scientific Research Program.}

\author[1,2]{Xingyu Xiao}[style=chinese]
\credit{Conceptualization, Methodology, Software, Formal analysis, Data Curation, Visualization, Validation, Writing- Original draft preparation}

\author[1,2]{Jiejuan Tong}[style=chinese]
\credit{Conceptualization, Formal analysis, Supervision, Writing - Review and Editing}

\author[1]{Jun Sun}[style=chinese]
\credit{Supervision}

\author[1]{Zhe Sui}[style=chinese]
\credit{Supervision}

\author[3]{Peng Chen}[style=chinese]
\credit{Software}

\author[1]{Jingang Liang}[orcid=0000-0003-2632-8613]
\cortext[cor1]{Corresponding author}
\cormark[1]
\ead{jingang@tsinghua.edu.cn; +86-10-62784836}
\credit{Resources, Supervision, Writing - Review and Editing, Project administration, Funding acquisition.}

\author[1]{Haitao Wang}[style=chinese]

\credit{Supervision, Writing- Reviewing and Editing}

\affiliation[1]{organization={Institute of Nuclear and New Energy Technology, Tsinghua University},
            city={Beijing},
            postcode={100084}, 
            country={China}}
\affiliation[2]{organization={National Key Laboratory of Human Factors Engineering},
            city={Beijing},
            postcode={100094}, 
            country={China}}

\affiliation[3]{organization={Institute of Software, Chinese Academy of Sciences},
city={Beijing},
postcode={100045}, 
country={China}}

\begin{abstract}
The rapid digitization of nuclear power plant main control rooms has fundamentally reshaped operator interaction patterns, introducing complex soft-control behaviors and elevated cognitive risks that are not adequately addressed by existing human reliability analysis approaches. Although recent advances in large language models and autonomous agents offer new opportunities for intelligent decision support, their deployment in safety-critical environments remains constrained by risks of hallucinated reasoning and weakened human authority. This study proposes NuHF-Claw, a persistent cognitive-risk agent framework that enables risk-governed human-centered autonomy for digital nuclear operations. The core methodological innovation lies in the introduction of a risk-constrained agent runtime, which tightly couples cognitive state inference with probabilistic safety assessment to regulate autonomous system behavior in real time. By integrating cognitively grounded workload and situational awareness estimation with dynamic human error probability prediction, the framework transforms conventional offline reliability analysis into a proactive intervention mechanism embedded directly within operational workflows. Experimental validation on a high-fidelity digital control room simulator demonstrates that NuHF-Claw can anticipate interface-induced cognitive degradation, dynamically constrain unsafe autonomous recommendations, and provide risk-aware navigational guidance while preserving human decision authority. The results highlight a fundamental shift from automation-driven operation toward cognition-aware autonomy, offering a principled pathway for the safe integration of intelligent agents into next-generation nuclear control environments.
\end{abstract}


\maketitle

\section{Introduction}
\label{Introduction}
The continuous evolution of nuclear power plant technology has fundamentally reshaped the design and operation of main control rooms (MCRs). This transformation has seen a shift from traditional analog instrumentation and control (I\&C) systems to sophisticated digital MCRs. While this transition to digital interfaces has been beneficial in improving information density, it presents a double-edged sword. On one hand, studies indicate a reduction in overall human error incidents, with reported percentages decreasing from 49\% to 41\%. On the other hand, this interface transformation has introduced complex operational behaviors, notably "soft control." Soft control describes the intricate process where operators actively navigate digital interfaces, employing various information processing strategies to collect plant status parameters and interpret system states. Crucially, this emergent behavior introduces novel modes of human error.

The advent of digital interfaces has been directly linked to the emergence of these new error types \cite{stubler2000soft}, which are predominantly cognitive in nature \cite{lee2011human}. These errors often manifest as challenges related to interface management, navigation complexities (e.g., mode confusion), and increased cognitive load. This highlights a critical insight: simply increasing system automation does not eliminate human error; rather, it shifts the operator's role to a supervisory one, altering the characteristics and underlying causes of potential failures. 

As the complexity of digital MCRs grows, human reliability assessment (HRA) methods have evolved to capture dynamic and cognitive characteristics, with frameworks like HUNTER \cite{boring2022hunter} representing significant advancements. However, current HRA methodologies remain largely passive and offline, struggling to provide real-time intervention for soft control behaviors. Concurrently, while large language models (LLMs) and autonomous AI agents offer transformative potential for real-time decision support, their direct application in safety-critical domains like nuclear power poses severe risks. Conventional AI agents lack a grounded understanding of human cognitive states and probabilistic risk constraints, leading to potential "hallucinations" and the dangerous erosion of human decision-making authority. Consequently, there is an urgent need for a paradigm shift: moving beyond passive dynamic HRA towards an active, risk-aware, and cognitively grounded agent architecture.

To address these critical limitations, we introduce \textbf{NuHF-Claw}, a persistent cognitive-risk agent framework designed for human-centered autonomy in digital MCRs. NuHF-Claw transforms static reliability evaluation into a "Risk-Constrained Agent Runtime." The framework integrates four core synergistic modules: 1) a Procedure-Interface Agent that maps parameter-seeking soft control tasks through a knowledge graph; 2) a Cognitive Twin Agent utilizing the ACT-R cognitive architecture to continuously monitor operator states in the background; 3) a Dynamic Risk Agent powered by a large language model (KRAIL) and grounded in the IDHEAS series framework for real-time performance influencing factor (PIF) analysis; and crucially, 4) a Governance Safety Gate that enforces human approval based on real-time risk thresholds. This work makes several significant contributions to the field of human-machine teaming and reliability engineering in digital MCRs:

\begin{itemize}
    \item \textbf{Addressing the "soft control" phenomenon through cognitive modeling}: This research systematically analyzes how soft control introduces novel, cognitive-heavy modes of human error. By explicitly integrating an ACT-R based Cognitive Twin Agent, the framework moves beyond traditional automation by actively simulating and predicting the operator's information processing bottlenecks and navigation complexities in real-time.
    \item \textbf{Establishing a Risk-Constrained Agent Architecture}: NuHF-Claw is introduced as a pioneering framework that bridges the gap between dynamic HRA and autonomous AI agents. By unifying simulator-driven data acquisition, knowledge graph-based task modeling, and KRAIL-based probabilistic reasoning, the architecture effectively restrains LLM-driven decision-making within strict, dynamically calculated human error probability (HEP) boundaries.
    \item \textbf{Enabling Human-Centered Autonomy and governed decision support}: Moving beyond static reliability assessments, this research lays the groundwork for safe AI integration in safety-critical systems. Through its Governance layer, NuHF-Claw ensures that autonomous suggestions are not only cognitively reasonable but probabilistically safe, effectively preventing AI overreach and strictly preserving the human operator's ultimate decision-making authority.
\end{itemize}

The remainder of this paper is structured as follows. Section \ref{Related Work} reviews the state of the art in human reliability analysis and related methodological streams. Section \ref{The NuHF-Claw Framework: An Integrated Approach} details the proposed framework and its underlying modeling principles. Section \ref{Case Study} demonstrates its applicability through a case study, covering scenario formulation, implementation methodology, empirical results, and result interpretation. Section \ref{Discussion} discusses the implications of the findings, emphasizing the strengths of the framework and benchmarking it against representative dynamic HRA approaches. Section \ref{Conclusion and future research} concludes the study and suggests avenues for future research.

\section{Related Work}
\label{Related Work}

\subsection{Human Reliability Analysis in Digital Environments} \label{Human Reliability Analysis}

Early human reliability analysis (HRA) methods, such as the technique for human error rate prediction (THERP), were primarily developed for analog control rooms and are categorized as "first-generation" techniques \cite{xiao2025dynamic}. These methods typically rely on a mechanistic, binary structure for error identification, offering limited analysis of underlying cognitive processes \cite{shirley2015validating, richei1996human}. Second-generation methods, such as ATHEANA and CREAM, shifted towards more theoretically grounded evaluations, explicitly accounting for cognitive errors and environmental conditions \cite{richei1996human}. However, both generations face profound limitations when applied to modern digital human-machine interfaces (HMIs). The protracted transition from analog to digital control rooms has caused traditional HRA methodologies to lag, often resulting in incomplete evaluations of digital-specific operator workloads \cite{boring2014human, porthin2016hra}. 

The modernization of nuclear power plants necessitates a fundamental adaptation of HRA methodologies \cite{porthin2020effects}. Recognizing that errors in "soft control" interfaces are frequently rooted in cognitive failures (e.g., mode confusion, information overload) rather than physical oversights, recent advancements have prioritized cognition-driven and empirically based approaches \cite{xiao2025novel}. The U.S. Nuclear Regulatory Commission's IDHEAS stands as a cutting-edge, technology-neutral method suitable for digital instrumentation and control (DI\&C) environments. It models human failure events as breakdowns in macro-cognitive functions (detection, understanding, decision-making, action execution) influenced by performance influencing factors (PIFs) \cite{xiao2024krail}. Similarly, methods like EMBRACE-APR1400 and HuRECA leverage empirical simulator data to evaluate response time reliability, cognitive task reliability, and design-related influencing factors specific to computerized functions \cite{kim2020hra, lee2012human}. 

Despite these strategic evolutions, a critical gap remains: contemporary HRA models are predominantly offline, retrospective, or rely on discrete simulator sessions. They lack the capacity for real-time, dynamic risk inference. Addressing the unique cognitive vulnerabilities introduced by soft controls requires a paradigm shift from static probability calculations to continuous, persistent cognitive-risk monitoring architectures.

\subsection{Cognitive Architectures and Human Digital Twins} \label{Cognitive Architectures in HRA}

To transition HRA from a descriptive to a predictive discipline, researchers increasingly integrate computational cognitive architectures. These architectures serve as both theoretical frameworks of the human mind and executable engines for simulation. Prominent models include GOMS-HRA, CMS-BN, IDAC, QN-MHP, and ACT-R.

GOMS-HRA adapts standard Keystroke-Level Model operators to automatically calculate human error probabilities based on contextual information \cite{boring2016goms}. However, it is primarily applicable to skilled users and inherently lacks deep mechanistic explanations for cognitive errors \cite{boring2018task}. CMS-BN utilizes Bayesian Networks to represent uncertain knowledge relationships but struggles to capture the continuous, fluid nature of real-time human-system interactions \cite{zhao2021cms}. IDAC and QN-MHP offer robust integrations with dynamic probabilistic risk assessment engines and multi-task simulations, respectively, though they can face accuracy challenges in predicting fine-grained task durations under specific interface geometries \cite{coyne2014nuclear, chen2024influence}.

Among these, ACT-R (Adaptive Control of Thought—Rational) stands out due to its profound mechanistic capabilities \cite{ritter2019act, lebiere1999dynamics}. ACT-R models errors stemming from fundamental cognitive limitations—such as memory decay, attentional shifts, and procedural reasoning—providing targeted insights rather than mere symptomatic classifications \cite{taatgen2006modeling}.  

Historically, a massive bottleneck in deploying ACT-R for nuclear HRA has been the immense complexity of manually authoring Lisp code to model intricate operator behaviors. Traditional manual coding is static, highly specialized, and struggles to scale to the dynamic variations of digital MCRs. However, recent advancements are overcoming this barrier by utilizing large language models to automatically generate ACT-R Lisp code, creating a highly automated toolchain. This paradigm effectively instantiates a persistent "Human Digital Twin." By automating the underlying Lisp execution, the model can dynamically simulate operator reaction times, a critical parameter for human reliability, and track cognitive trajectories in real-time, providing the essential algorithmic foundation for proactive risk agents.

\subsection{Procedural Task Modeling for Autonomous Systems}
\label{Procedural Task Modeling in HRA}

Task modeling dissects complex operational goals into identifiable steps, allowing for a precise evaluation of human-system interactions. First-generation HRA relied on sequential, behavior-oriented decomposition, treating operators akin to machine components \cite{xiao2025autograph}. Second-generation methods integrated the Stimulus-Organism-Response (S-O-R) paradigm, acknowledging the internal cognitive states and specific environmental contexts driving user actions \cite{baumberger2005cooperative, hollnagel1998cognitive}.

Recent innovations, such as the Hunter framework, push task modeling further by coupling a "virtual operator" with a simulated nuclear system to generate predictive empirical data \cite{boring2022hunter}. However, frameworks like Hunter face significant practical obstacles: high entry barriers requiring specialized modeling languages and limited generalization capabilities for new emergency operating procedures (EOPs). 

For modern autonomous agents to assist in MCRs, task modeling must evolve into machine-readable procedural graphs \cite{xiao2026intelligent}. There is an urgent necessity for automated methodologies capable of parsing natural language protocols, decomposing them into fine-grained atomic actions, and semantically mapping them to digital interface elements. Such graph-based representations are imperative for agents to execute "procedure comprehension" and verify execution paths without human modeling bottlenecks.

\subsection{Dynamic Performance Influencing Factors Analysis}

Performance Influencing Factors (PIFs), or Performance Shaping Factors (PSFs), are the critical variables that modulate human reliability. While static PIFs (e.g., fixed equipment labels) are easily audited, dynamic PIFs (e.g., operator cognitive workload, dynamic scenario complexity, and temporal stress) fluctuate continuously and require real-time data streams for accurate assessment \cite{xiao2025dynamic}.

Traditional methods, such as SPAR-H, rely on static multipliers or simplified dynamic formulas (e.g., calculating complexity based on discrete plant trip states) \cite{gertman2005spar}. While objective, these simplified correlations are insufficient for capturing the nuanced, interface-induced errors prevalent in soft control environments. The IDHEAS-ECA methodology provides a more comprehensive, cognition-grounded approach to PIFs, yet it demands substantial analytical effort.

To bridge this gap, frameworks like KRAIL have emerged, integrating the comprehensive IDHEAS database with the computational reasoning of LLMs \cite{xiao2024krail}. By utilizing multi-agent systems and knowledge graphs, KRAIL semi-automates the estimation of baseline human error probabilities, significantly enhancing efficiency. However, to achieve true human-centered autonomy, this LLM-driven PIF analysis must transcend static prompt-response mechanisms. It must be embedded within a continuously running agent runtime, where dynamic PIFs continuously shape the probability thresholds of safety gates, ensuring that agent recommendations remain strictly risk-constrained.

\subsection{Autonomous Agents in Safety-Critical Systems}
\label{Autonomous Agents in Safety-Critical Systems}

Recent advancements in artificial intelligence have catalyzed a paradigm shift from simple conversational LLMs to persistent \cite{wang2024survey}, autonomous agents. These agent frameworks are characterized by their ability to maintain long-term memory, invoke external tools, and execute complex, multi-step planning loops (e.g., observe-assess-act) \cite{miller2021multi}. In general software and commercial domains, these autonomous processes are highly effective. However, directly transferring these unconstrained agent runtimes to safety-critical systems, such as nuclear power plant (NPP) digital main control rooms (MCRs) \cite{park2017experimental}, introduces profound operational risks. 

The primary challenge lies in the inherent stochastic nature of LLMs. In highly regulated and unforgiving environments, phenomena such as "hallucinations, where an AI generates plausible but factually incorrect or unsafe operational steps—pose an unacceptable threat.  Furthermore, granting unconstrained execution authority to an AI agent fundamentally violates the defense-in-depth principle and erodes the human operator's ultimate decision-making authority \cite{xiao2025integrating}. Most existing agent architectures are designed for task completion rather than risk containment, lacking the strict, probabilistic safety boundaries required for nuclear operations.

To mitigate these risks and safely deploy autonomous agents in NPPs, there is an emerging consensus that AI reasoning must be tightly coupled with deterministic engineering models and established human cognitive architectures \cite{kumar2025cognitive}. Traditionally, developing robust cognitive models like ACT-R to simulate operator behavior has been a major bottleneck. It requires the complex, manual authoring of Lisp code, a process that is not only labor-intensive but also results in static models that struggle to adapt to the real-time, dynamic nature of digital HMIs. 

However, the advanced code-generation and reasoning capabilities of modern LLMs present a transformative solution to this bottleneck. Instead of acting as direct decision-makers, LLMs can be utilized as automated orchestration tools to dynamically generate and manipulate ACT-R Lisp code based on real-time plant telemetry and interface interaction logs \cite{yang2025code}. This synergy enables the creation of a continuously running "Human Digital Twin." By automating the instantiation of the cognitive model, the system can dynamically simulate human reaction times, a critical parameter for human reliability analysis, and track the operator's mental trajectory in real-time, identifying potential mode confusion or cognitive overload before an error occurs.

Despite these theoretical opportunities, a comprehensive framework that unifies persistent AI agents, automated cognitive twin generation, and dynamic probabilistic risk assessment remains absent in the nuclear domain \cite{ruscio2017distraction}. The industry urgently requires a transition from simplistic "operator replacement" concepts toward "human-centered autonomy." This necessitates a novel risk-constrained agent runtime: an architecture where an agent's recommendations are continuously governed by real-time cognitive state estimations and strict safety gates, ensuring that the AI acts as an advanced cognitive co-pilot rather than an autonomous dictator. The NuHF-Claw framework proposed in this paper is designed to fill this exact methodological gap.

\section{The NuHF-Claw Framework: An Integrated Approach}
\label{The NuHF-Claw Framework: An Integrated Approach}

\subsection{Overview of NuHF-Claw: A Risk-Constrained Agent Runtime}
\begin{figure}[h]
\centering
\includegraphics[width=0.9 \textwidth]{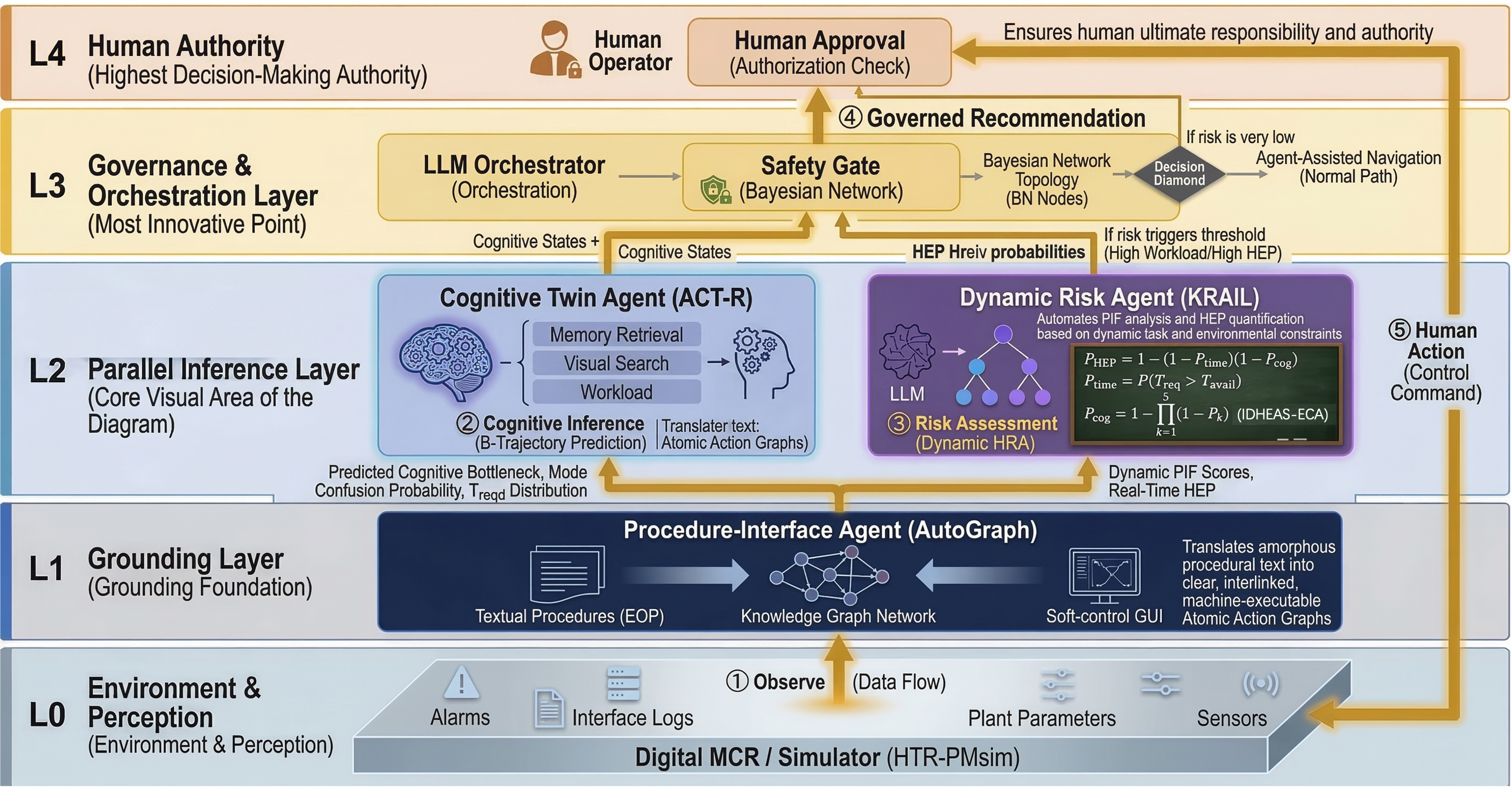}
\caption{The Architecture of NuHF-Claw: A Risk-Constrained Multi-Agent Runtime}\label{framework}
\end{figure}

Unlike traditional, passive dynamic HRA methodologies that focus solely on post-event or offline probability calculations, NuHF-Claw is engineered as a persistent, multi-agent runtime. It achieves human-centered autonomy through a sophisticated, interconnected workflow among four specialized cognitive and risk agents. This synergistic architecture, illustrated in Figure \ref{framework}, operates continuously in the background of digital MCRs, functioning as a safety-critical copilot rather than an unconstrained autonomous controller.

The runtime initiates with the \textbf{MCR Digital Twin Environment} (simulator-driven data acquisition), which continuously monitors system states and operator interactions. When an initial event is detected, the workflow activates the \textbf{Procedure-Interface Agent} (integrating knowledge graph methodologies). This agent parses textual emergency operating procedures (EOPs) and maps them to the real-time soft-control interface, grounding potential actions in a deterministic graph to prevent AI hallucinations. 

Simultaneously, the proposed operational paths are fed into two parallel evaluation agents. The \textbf{Cognitive Twin Agent} employs the ACT-R architecture to simulate the human operator's cognitive trajectory, predicting time requirements and potential mode confusion. Concurrently, the \textbf{Dynamic Risk Agent} (powered by the KRAIL large language model framework) calculates real-time PIFs and contextual human error probabilities ($P_c$) based on IDHEAS-ECA theory.

Finally, instead of directly executing actions, these multidimensional assessments converge at the \textbf{Governance Safety Gate}. This novel component utilizes probabilistic inference (e.g., Bayesian Networks) to weigh the predicted cognitive load against the dynamic risk score. If the risk exceeds predefined safety thresholds, the system deliberately restricts autonomous action, generating a risk-informed suggestion and enforcing human approval. This seamless integration forms a truly dynamic, risk-constrained mechanism for human-machine teaming.

\subsection{Key Agent Components}

\subsubsection{MCR Digital Twin Environment (Simulator Data Acquisition)}
The foundation of the NuHF-Claw runtime relies on continuous environmental perception. The framework leverages the dynamic risk-informed framework (DRIF) \cite{xiao2025dynamic} to establish a digital twin of the MCR. This study utilizes the HTR-PMsim simulator, specifically developed and validated for the high-temperature gas-cooled reactor demonstration plant (HTR-PM). Real-time operational time-series data (comprising 33 operational parameters across various scenarios) is continuously fed into an LSTM-based classification algorithm within the DRIF module to automatically diagnose initial events. This eliminates human errors associated with manual fault diagnosis and triggers the subsequent agent workflows.

\subsubsection{Procedure-Interface Agent (Task Modeling via AutoGraph)}\label{Task Modeling}
To bridge the semantic gap between textual procedures and dynamic interfaces, NuHF-Claw deploys a Procedure-Interface Agent powered by the AutoGraph framework \cite{xiao2025autograph}. AutoGraph establishes a structured foundation for semantic modeling of soft-control interactions. Unobtrusive trackers capture operator actions to construct an interface-embedded knowledge graph (IE-KG). Procedural steps are then explicitly mapped to execution paths within this IE-KG. By flagging multi-action steps and resolving interface hierarchies, this agent translates amorphous procedural text into clear, interlinked, and machine-executable operational paths, acting as a deterministic constraint against LLM deviations.

\subsubsection{Cognitive Twin Agent (ACT-R based Human Modeling)}
Rather than solely calculating static execution times, NuHF-Claw introduces a persistent Cognitive Twin Agent. Leveraging the ACT-R (adaptive control of thought—rational) cognitive architecture \cite{xiao2025cognitive}, this agent operates in parallel with the human operator. It automatically ingests the operational paths generated by the Procedure-Interface Agent to simulate the operator's cognitive processes (e.g., visual search on soft controls, memory retrieval, and motor execution). 

This simulation facilitates the estimation of required time ($T_{reqd}$). Following NRC guidance (NUREG-2256, RIL 2024-17) \cite{xing2022integrated, HunterRIL}, the agent models the $T_{reqd}$ distribution as lognormal, utilizing a recommended shape parameter ($\sigma = 0.28$) and calculating the scale parameter ($\mu$) based on conservative point estimates. This generates a dynamic distribution of time-related human error probability ($P_t$), serving as a real-time indicator of operator cognitive bottlenecks.

\subsubsection{Dynamic Risk Agent (PIF Analysis via KRAIL)}
To evaluate contextual risks, the framework employs a Dynamic Risk Agent driven by KRAIL (Knowledge-based reliability analysis using an intelligent large language model) \cite{xiao2025krail}. Grounded in the IDHEAS-ECA methodology, KRAIL overcomes the subjectivity of traditional expert judgment. The agent operates a two-stage process: first, a multi-agent prompt framework decomposes the task context; second, an integration framework calculates the base HEP and PIF attributes via few-shot learning and KG injection. Crucially, KRAIL ensures efficiency—processing complex dynamic scenarios in under 150 seconds—thereby providing the runtime with continuous, real-time probabilistic constraints ($P_c$) reflecting current environmental complexities.

\subsubsection{Governance Safety Gate (Bayesian Inference Layer)}\label{Safety Gate}
To resolve the critical challenge of AI overreach, NuHF-Claw introduces the Governance Safety Gate. \textit{(Here we answer the Bayesian Network question)} This layer acts as the ultimate authority preserver. It employs a Bayesian Network (BN) to synthesize the outputs from the Cognitive Twin Agent ($P_t$, workload, confusion state) and the Dynamic Risk Agent ($P_c$, PIF severity). The BN infers a holistic "Action Risk Probability." 

Instead of allowing the LLM to directly execute a system command, the Safety Gate compares this probability against predefined safety margins. If the inferred risk is low, it may allow minor interface navigation assistance. However, if cognitive overload or high HEP is detected, the Gate restricts the system to generating "Action Suggestions" accompanied by risk explanations, explicitly mandating human operator approval before execution. This ensures the autonomy remains strictly human-centered.

\subsection{Architectural Design Principles}
The engineering of NuHF-Claw is guided by four principles tailored for AI integration in safety-critical domains:

\begin{itemize}
    \item \textbf{Explainability and Auditability (Interpretability)}: The runtime must provide transparent insights. By grounding LLM reasoning in the AutoGraph knowledge graph and IDHEAS-ECA theory, the framework explains not just "what" the suggested action is, but "why" it is recommended based on dynamic PIFs, ensuring every agent intervention is fully auditable.
    \item \textbf{Human-Centered Autonomy (Evolution of Automation)}: Rather than replacing the operator to achieve full automation, NuHF-Claw is designed to augment human capability. It automates the cognitive heavy-lifting (e.g., PIF assessment and procedure tracing) while utilizing the Governance Safety Gate to preserve the operator's final decision authority.
    \item \textbf{Agent Isolation (Modularity)}: The system is structured with independent, interchangeable agent layers (Perception, Cognition, Risk, Governance). This prevents a single LLM from handling end-to-end tasks, thereby isolating potential "hallucination" faults and allowing seamless integration of future mechanism models or advanced reactor data.
    \item \textbf{Persistent Runtime (Real-time Capability)}: Moving beyond static, post-event HRA tools, NuHF-Claw operates continuously. It adapts to evolving system configurations and human states on the fly, offering proactive, risk-informed navigational support and dynamic intervention in live digital control room settings.
\end{itemize}

\section{Case Study: Validating the NuHF-Claw Framework}
\label{Case Study}

To empirically validate the NuHF-Claw framework, we conducted a comprehensive case study using the HTR-PM600 (High-Temperature Gas-Cooled Reactor) full-scope simulator. The study demonstrates how NuHF-Claw operates as a persistent, risk-constrained agent to monitor, assess, and support operators during a critical transient event.

\subsection{Scenario Description and Experimental Setup}

\begin{figure}[h]
\centering
\includegraphics[width=0.6\textwidth]{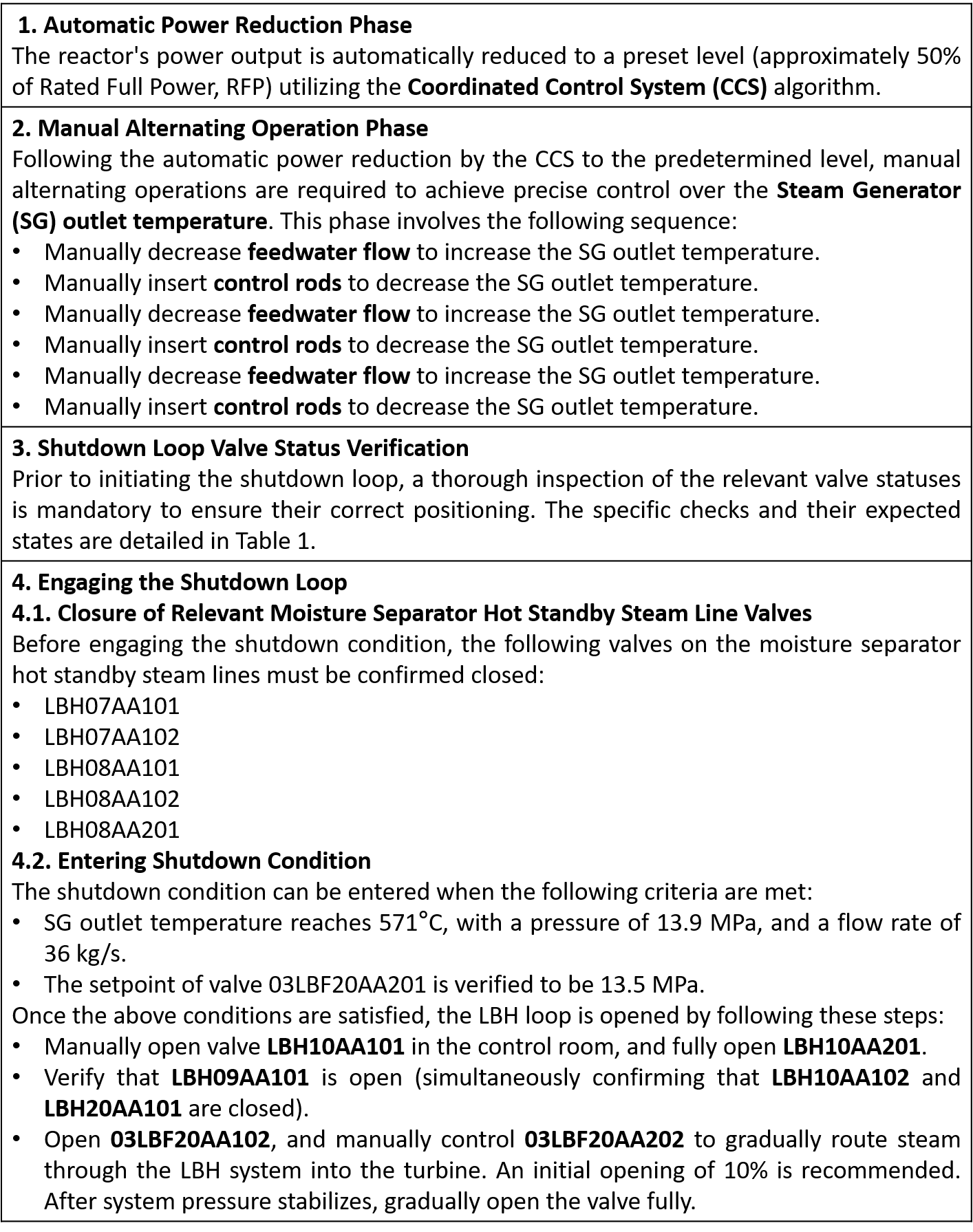}
\caption{Reactor Shutdown Procedure Mapped onto the Interface-Element Knowledge Graph (IE-KG)}\label{procedure}
\end{figure}

The operational scenario investigated in this study is illustrated in Figure \ref{procedure}. Specifically, the experimental setting is based on a representative set of reactor shutdown test procedures, including automatic power reduction, manual alternating control of steam generator temperature, shutdown loop status verification, and shutdown condition engagement. These procedures constitute a typical structured operational sequence in digital nuclear control rooms and provide a realistic context for evaluating interface-driven cognitive and behavioral responses.

\begin{figure}[h]
\centering
\includegraphics[width=0.6\textwidth]{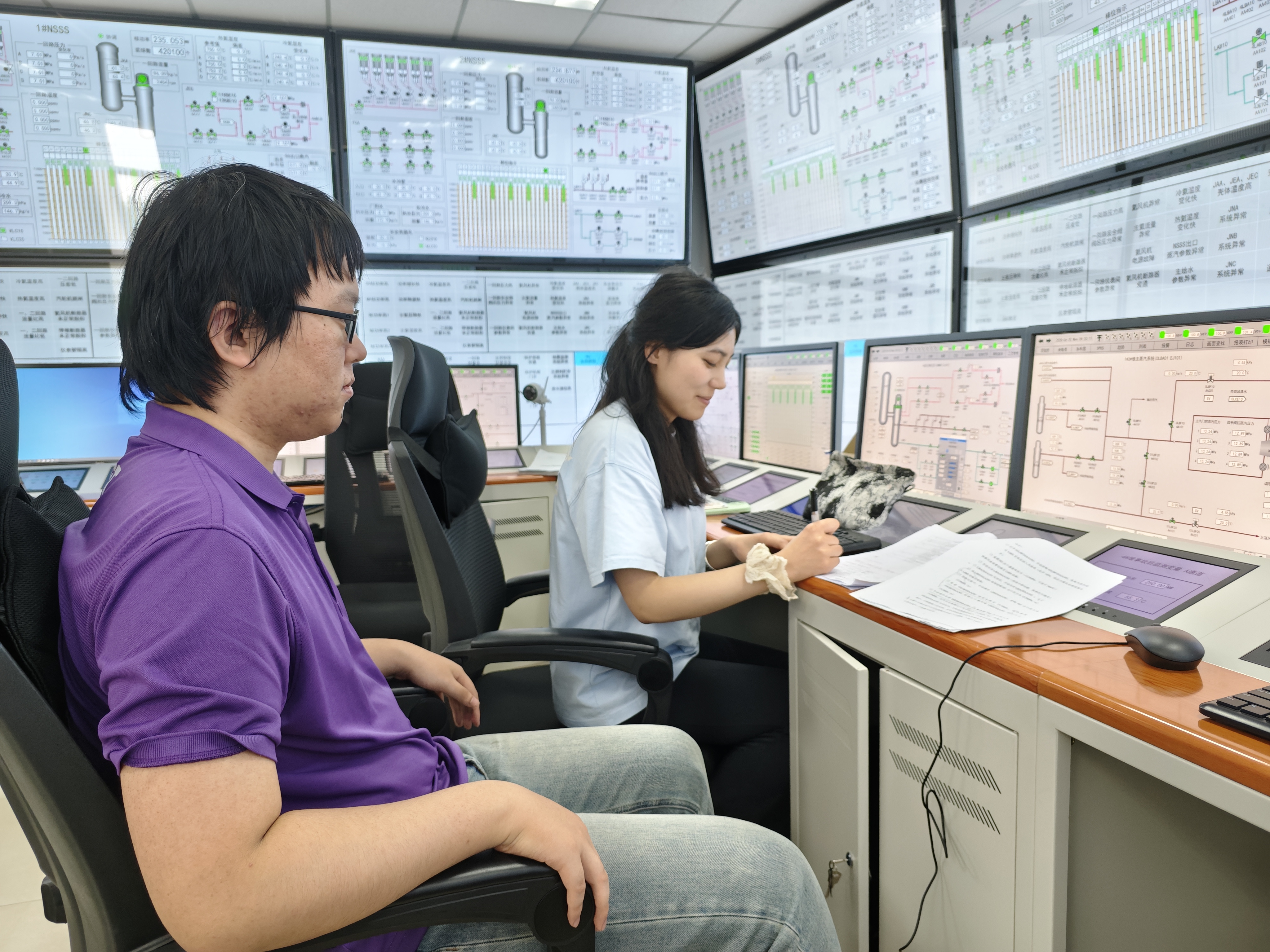}
\caption{Students Conducting Experiments on the HTR-PM600 1:1 Full-Scope Simulator}\label{HTGRmiao}
\end{figure}

The study recruited three graduate students majoring in nuclear science and technology (two master's and one doctoral candidate). Comprehensive training was provided to ensure familiarity with the simulator (Figure \ref{HTGRmiao}). To induce realistic cognitive load and operational stress, replicating emergency conditions, a performance-based time incentive was introduced, encouraging participants to complete tasks rapidly. Each participant spent 20 to 40 minutes, yielding approximately 3 hours of operational data.

\begin{figure}[h]
\centering
\includegraphics[width=0.6\textwidth]{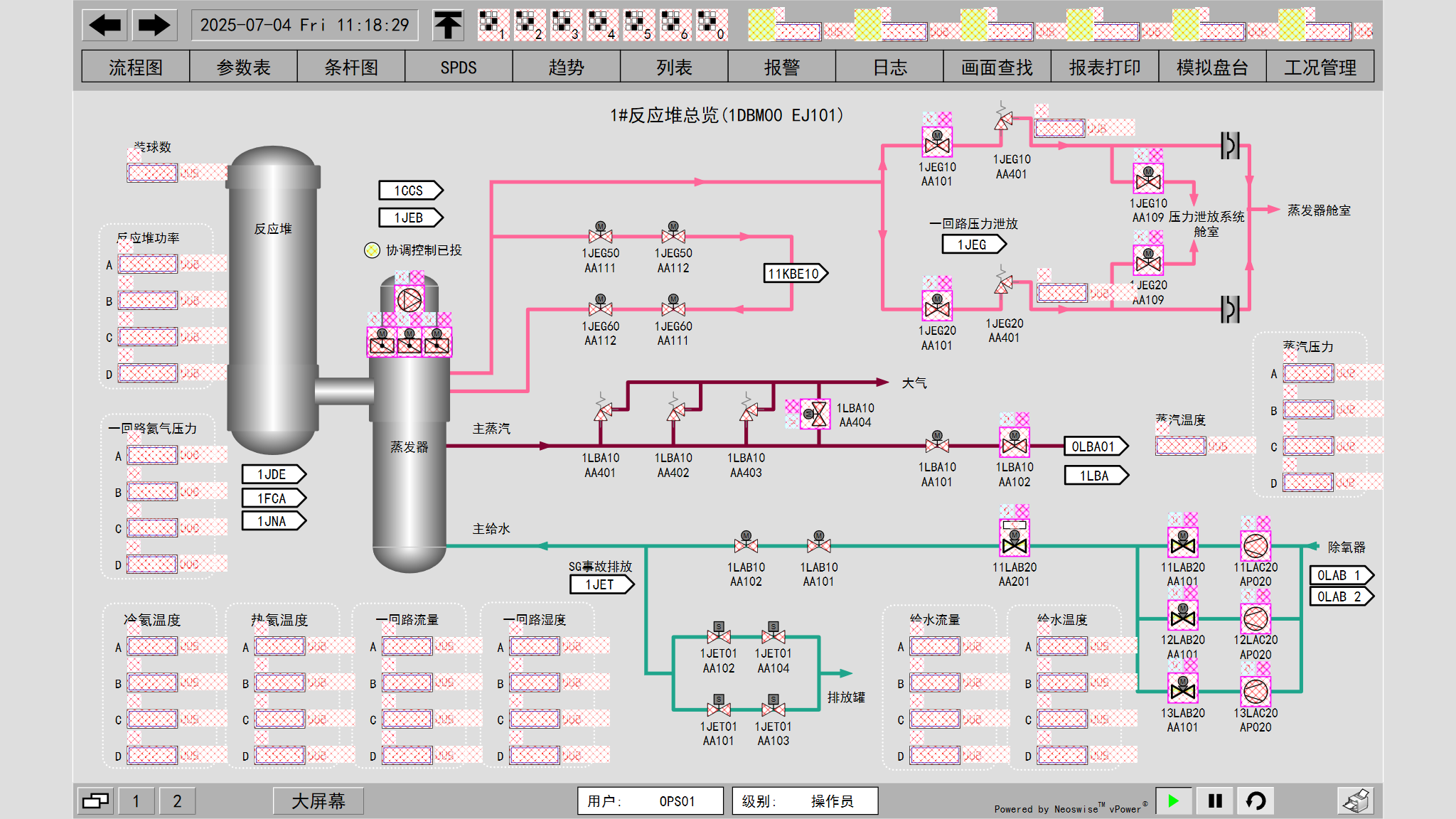}
\caption{Interface-Element Knowledge Graph (IE-KG) Derived from the HTR-PM600 Interface}\label{simulator}
\end{figure}

Three streams of multimodal data were continuously collected to feed the NuHF-Claw Perception Layer: (1) full-screen recordings of the simulator interface (Figure \ref{simulator}); (2) real-time cursor trajectories and click coordinates via a custom logging tool (\textit{tracker.exe}); and (3) external video footage for observational behavioral analysis. 

To reflect realistic operational governance, participants employed a human performance tool (HPT), the "two-step verification card." Participants circled intended actions during the reading phase and marked an "X" post-execution. This physical governance effectively eliminated Errors of Omission (EOO) during the baseline sessions, providing a clean dataset for evaluating Errors of Commission (EOC) and interface-induced cognitive delays.

\begin{table}[width=.9\linewidth,cols=5,pos=h]
\caption{Shutdown Loop Valve Status Checklist}\label{Checklist}%
\begin{tabular*}{\tblwidth}{@{} LLLLL@{} }
\toprule
\textbf{No.} & \textbf{Valve Code} & \textbf{Valve Name} & \textbf{Expected State} &  \textbf{Actual} \\
\midrule
1  & LBH10AA101     & Moisture Separator Inlet Isolation Valve 1   & Closed & [    ]  \\
2  & LBH10AA201     & Moisture Separator Inlet Regulating Valve   & Open   & [    ] \\
3  & LBH10AA102     & Moisture Separator Inlet Isolation Valve 2   & Closed &[    ] \\
4  & LBH20AA101     & Moisture Separator Outlet Isolation Valve  & Closed &[    ] \\
5  & LBH09AA101     & Moisture Separator Bypass Isolation Valve  & Open   &  [    ]\\
6  & LBH30AA101     & Moisture Separator Drain Isolation Valve   & Open  &  [    ] \\
7  & LBH30AA201     & Moisture Separator Drain Regulating Valve   & Open   &  [    ] \\
8  & LBH50AA101     & Moisture Separator Drain to Condenser Isolation Valve & Open  &  [    ] \\
9  & LBF20AA201   & Reactor Turbine Bypass Valve    & Auto  &  [    ] \\
10 & LBF20AA101   & Reactor Main Steam to Bypass Motorized Isolation Valve   & Open  &  [    ] \\
11 & LBA20AA101     & Reactor Main Steam Motorized Isolation Valve 1  & Open  &  [    ] \\
12 & LBA20AA102     & Reactor Main Steam Motorized Isolation Valve 2     & Open  &  [    ] \\
\bottomrule
\end{tabular*}
\end{table}

\subsection{NuHF-Claw Methodology Application}

The application of NuHF-Claw follows its four-layer architecture. First, the \textbf{Perception Layer} executes simulator-driven data acquisition and state identification. We utilized a Long Short-Term Memory (LSTM) network to ingest the continuous telemetry data from the HTR-PM600 simulator (e.g., thermal power, helium blower speed, condenser water level, as shown in Table \ref{Standardized}). The LSTM successfully identified the initial transient event as the "Disconnection of Generator to 6kV 1B Bus bar."

\begin{figure}[h]
\centering
\includegraphics[width=0.6\textwidth]{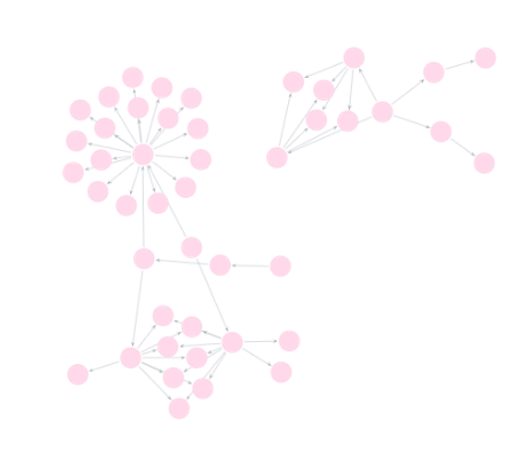}
\caption{Interface-Element Knowledge Graph (IE-KG) Derived from the HTR-PM600 Interface}\label{KG}
\end{figure}

\begin{figure}[h]
\centering
\includegraphics[width=0.6\textwidth]{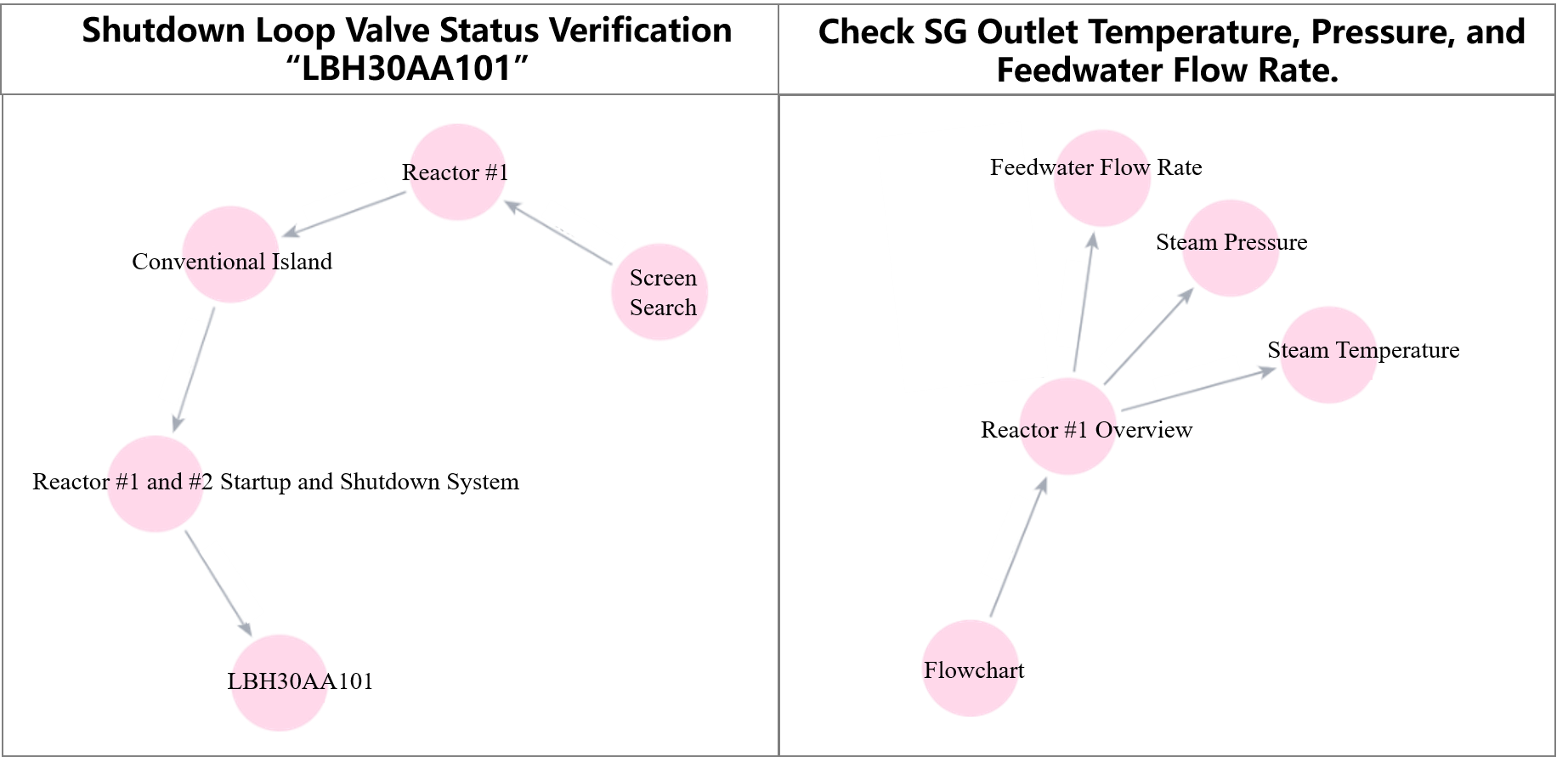}
\caption{AutoGraph Mapping of Textual Procedures to Interface Coordinates}\label{sample}
\end{figure}

\begin{figure}[h]
\centering
\includegraphics[width=0.6\textwidth]{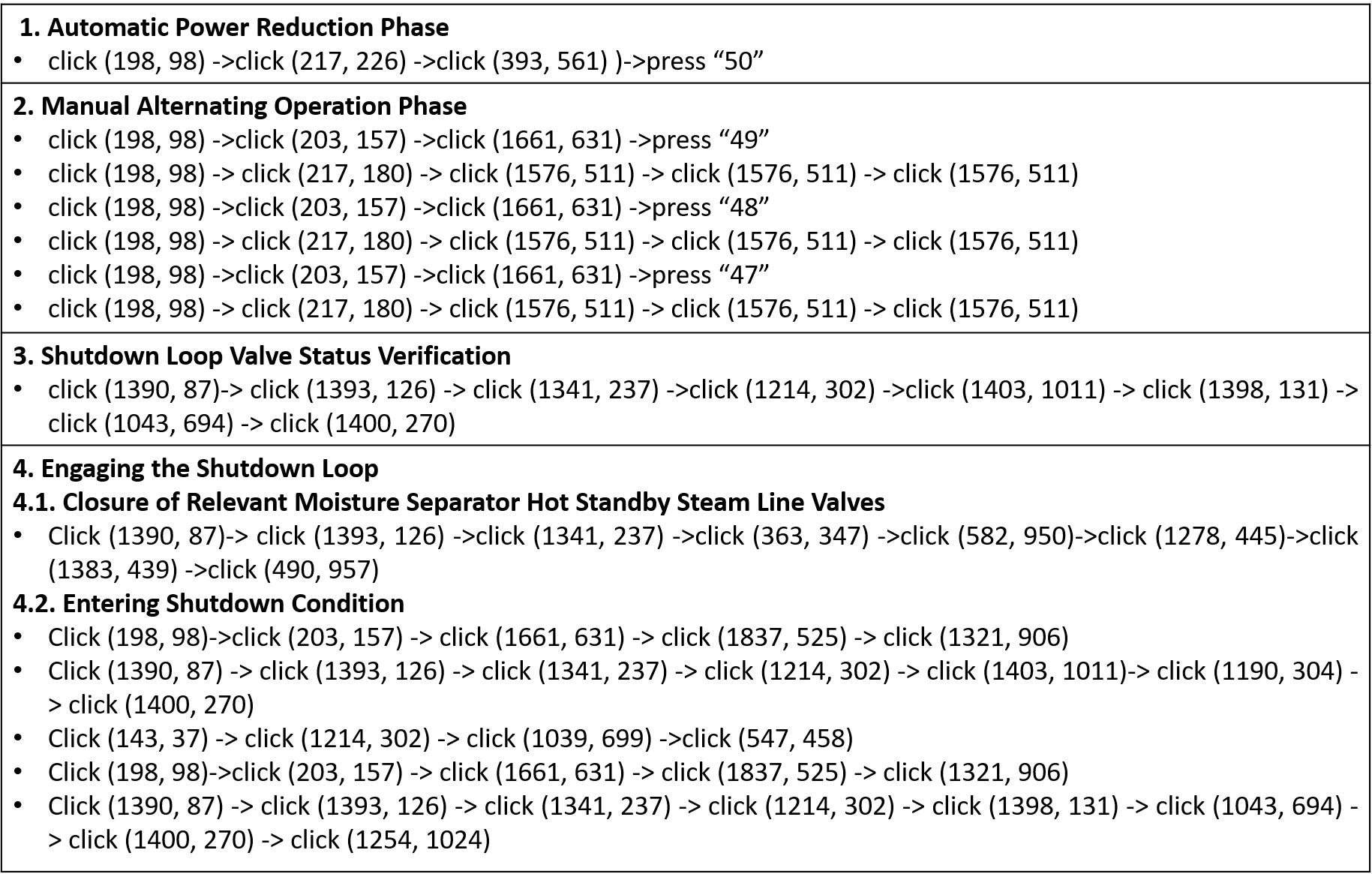}
\caption{Complete Executable Graph Representation for Reactor Shutdown}\label{procedure2}
\end{figure}

Second, the \textbf{Procedure-Interface Agent} parsed the corresponding Emergency Operating Procedures (EOPs). Using the AutoGraph framework, textual procedures were automatically mapped onto an Interface-Element Knowledge Graph (IE-KG) representing the HTR-PM600 soft-control panels. Figure \ref{KG} illustrates the hierarchical structure of this graph, where green nodes denote top-level elements and pink nodes represent deeper navigation layers. This dynamic mapping transforms static text into executable, coordinate-grounded operational paths (as visualized in Figure \ref{sample} and Figure \ref{procedure2}).

Third, the \textbf{Cognitive Twin Agent} instantiated a persistent ACT-R model based on the AutoGraph-generated paths. By automatically generating Lisp code representing the operator's mental trajectory, the twin simulated the continuous cognitive processes (perception, memory retrieval, action), yielding the precise Time Required ($T_{reqd}$) distribution for the operations. This output forms the basis for calculating the time-dependent failure probability ($P_t$). 

Finally, the \textbf{Dynamic Risk Agent} employed the KRAIL framework, leveraging an LLM to assess real-time PIFs based on IDHEAS macro-cognitive theory. This produced the cognitive failure probability ($P_c$). The \textbf{Governance Layer} then utilized a Bayesian Network to dynamically couple $P_t$ and $P_c$, mapping the multi-step procedures into a holistic, real-time systemic Human Error Probability (HEP) metric, dictating whether autonomous intervention or human confirmation is required.

\begin{table}[htbp]
\centering
\caption{Standardized Telemetry Data Sample from the HTR-PM600 Simulator.}
\label{Standardized}
\resizebox{\linewidth}{!}{
\begin{tabular}{ccccccccc}
\toprule
\textbf{TIME (10 ms)} & 
\textbf{Nuclear Power} & 
\textbf{Thermal Power \#1} & 
\textbf{Helium Blower Speed \#1} & 
\textbf{...} & 
\textbf{Hot Helium Temp \#2} & 
\textbf{Cold Helium Temp \#2} & 
\textbf{...} & 
\textbf{Condenser Level} \\
\midrule
17    & 183.5995 & 0        & 3826.837 & $\dots$ & 542.2744 & 251.3571 & $\dots$ & 704.1935 \\
37    & 183.6040 & 198.2936 & 3820.901 & $\dots$ & 542.2758 & 251.3527 & $\dots$ & 704.2556 \\
10610 & 183.0486 & 199.7107 & 3800 & $\dots$ & 519.5561 & 248.2146 & $\dots$ & 603.6227 \\
\bottomrule
\end{tabular}
}
\end{table}

\subsection{Results}

\textbf{Workload Perception and Cognitive Twin Alignment:}
The HTR-PM600 employs a multi-module operation concept where one control room manages two reactors, involving multiple roles (e.g., 1RO, 2RO, 3RO). The NuHF-Claw Cognitive Twin Agent continuously estimated the real-time workload for these roles. As detailed in Table \ref{Workload}, the estimations revealed profound disparities. 1RO bore the brunt of the emergency response, exhibiting critical peaks in mental workload (100), temporal demand (100), and effort (100), resulting in an aggregated TLX-equivalent score of 55.83. Conversely, 2RO and 3RO demonstrated significantly lower temporal demand (20) and overall scores (33.33 and 36.67, respectively). The model's predictive performance demonstrated exceptional stability (Explained Variance > 0.9, $R^2 > 0.9$), confirming the Cognitive Twin's capacity to precisely track distributed operator fatigue during complex transients.

\begin{table}[width=.9\linewidth,cols=6,pos=h]
\caption{Cognitive Twin Predicted Workload Distribution Across Operator Roles}\label{Workload}%
\begin{tabular*}{\tblwidth}{@{} LLLL@{} }
\toprule
\textbf{Dimension} & \textbf{1RO} & \textbf{2RO} & \textbf{3RO}  \\
\hline
Mental Workload   & 100 & 70  & 80 \\
Physical Demand   & 10  & 10  & 10 \\
Temporal Demand   & 100 & 20  & 20 \\
Performance       & 95  & 80  & 80 \\
Effort            & 100 & 70  & 80  \\
Frustration       & 20  & 10  & 10\\
\hline
\textbf{Aggregated Score}      & \textbf{55.83} & \textbf{33.33} & \textbf{36.67}  \\
\hline
\end{tabular*}
\end{table}

\textbf{Automated Procedural Mapping and Execution Time Inference:}
Following the initial event detection, the AutoGraph module successfully mapped the "Reactor Shutdown" procedures. Figure \ref{sample} illustrates the precise alignment of textual directives (e.g., "Check SG Outlet Temperature") to localized screen coordinates. The complete categorized operational paths (summarized in Table \ref{procedure3}) reveal the heavy reliance on complex "Screen Navigation" and "Top-Left Toggle" behaviors inherent to digital HMIs.

By feeding these precise paths into the ACT-R based Cognitive Twin, NuHF-Claw generated continuous execution time distributions ($T_{reqd}$). According to NUREG-2256 guidelines, for in-control-room actions, a log-normal distribution with a shape parameter ($\sigma$) of 0.28 was adopted. The scale parameter ($\mu$) was dynamically derived using the median times generated by the ACT-R Lisp execution, thereby fully automating the derivation of $P_t$. Concurrently, the KRAIL module reasoned over these paths to compute the cognitive error probability ($P_c$), successfully fusing them within the Bayesian Governance layer to output a continuous risk trajectory.

\subsection{Discussion of Results}

\textbf{Empirical Validation of the Cognitive-Risk Agent:}
The results demonstrate the validity of NuHF-Claw's predictive capabilities. The $T_{reqd}$ distributions generated dynamically by the ACT-R Lisp code tightly aligned with the empirical cursor-tracking data (\textit{tracker.exe}) collected from the six participants. This alignment validates the premise that automating ACT-R instantiation via LLMs can effectively function as a real-time Human Digital Twin. Furthermore, the workload estimates computed by the Cognitive Twin were corroborated by post-experiment NASA-TLX questionnaires administered to the operators, confirming the model's accuracy in capturing real-world cognitive bottlenecks.

\textbf{Shifting to Human-Centered Autonomy:}
Unlike traditional static HRA methods, NuHF-Claw provides proactive insights. Table \ref{procedure3} highlights that sequences like "Screen Navigation 1" (requiring 11 consecutive interactions) introduce significant cognitive vulnerabilities due to mode confusion and memory decay. Rather than passively recording this risk or autonomously taking over the plant, NuHF-Claw's Governance Layer is designed to flag these specific high-risk navigational nodes, dynamically suggesting procedural shortcuts or enforcing strict human-confirmation gates. This explicitly preserves the human's final decision authority while mitigating the digital interface's inherent risks.

\begin{table}[width=.9\linewidth,pos=h]
\scriptsize
\caption{Categorized Operational Paths for Procedure Execution in Figure \ref{procedure2}}\label{procedure3}%
\begin{tabular*}{\tblwidth}{@{} p{0.25\linewidth} p{0.75\linewidth} @{}}

\toprule
\textbf{Category} & \textbf{Operational Paths/Control Points} \\
\hline
\multirow{5}{*}{Flowchart Execution 1} & Procedure (198, 95) \\
& --- Coordination Control (217, 226) \\
& --- Thermal Power Setpoint (393, 561) \\
& --- Parameter Tuning \\
& --- Parameter Tuning End \\
\hline
\multirow{3}{*}{Flowchart Execution 2} & Procedure (198, 95) \\
& --- Reactor Power Control (217, 180) \\
& --- Rod Insertion (1576, 511) \\
\hline
\multirow{5}{*}{Flowchart Execution 3} & Procedure (198, 95) \\
& --- Reactor Overview (209, 155) \\
& --- Steam Temperature (1661, 631) \\
& --- Steam Pressure (1839, 525) \\
& --- Feedwater Flow (1321, 506) \\
\hline
\multirow{11}{*}{Screen Navigation 1} & Screen Lookup (1390, 87) \\
& --- Reactor (1393, 126) \\
& --- Conventional Island (1341, 237) \\
& --- I\#2\# Startup/Shutdown System (363, 347) \\
& --- LBH0AA101 (484, 333) \\
& --- LBH0AA201 (574, 327) \\
& --- LBH0AA102 (761, 326) \\
& --- LBH20AA101 (1383, 324) \\
& --- LBH0AA103 (1025, 234) \\
& --- LBH20AA101 (1157, 570) \\
& --- LBH30AA201 (1273, 371) \\
& --- LBH50AA101 (1508, 577) \\
\hline
\multirow{8}{*}{Screen Navigation 2} & Screen Lookup (1390, 87) \\
& --- Reactor (1393, 126) \\
& --- Conventional Island (1341, 237) \\
& --- I\#2\# Main Steam System (1214, 302) \\
& --- LBF20AA201 (1190, 304) \\
& --- LBF20AA101 (1183, 310) \\
& --- LBA20AA101 (1398, 131) \\
& --- LBA20AA102 (1331, 236) \\
\hline
\multirow{8}{*}{Screen Navigation 3} & Screen Lookup (1390, 87) \\
& --- Reactor (1393, 126) \\
& --- Conventional Island (1341, 237) \\
& --- I\#2\# Startup/Shutdown System (363, 347) \\
& --- LBH07AA101 (490, 957) \\
& --- LBH07AA102 (582, 950) \\
& --- LBH08AA101 (378, 442) \\
& --- LBH08AA102 (1383, 439) \\
\hline
\multirow{5}{*}{Screen Navigation 4} & Screen Lookup (1390, 87) \\
& --- Reactor (1393, 126) \\
& --- Conventional Island (1341, 237) \\
& --- I\#2\# Main Steam System (1214, 302) \\
& --- LBF20AA201 (1190, 304) \\
\hline
\multirow{8}{*}{Screen Navigation 5} & Screen Lookup (1390, 87) \\
& --- Reactor (1393, 126) \\
& --- Conventional Island (1341, 237) \\
& --- I\#2\# Reactor Main Steam System (1214, 302) \\
& --- LBA20AA101 (1398, 131) \\
& --- LBA20AA102 (1331, 236) \\
& --- LBF20AA101 (1190, 304) \\
& --- LBF20AA202 (647, 458) \\
\hline
\multirow{7}{*}{Top-Left Toggle 1} & Top-Left Toggle (143, 37) \\
& --- I\#3\# Startup/Shutdown System (363, 347) \\
& --- LBH0AA101 (484, 333) \\
& --- LBH10AA201 (574, 327) \\
& --- LBH09AA101 (1025, 234) \\
& --- LBH20AA102 (761, 326) \\
& --- LBH20AA201 (1383, 324) \\
\hline
\multirow{4}{*}{Top-Left Toggle 2} & Top-Left Toggle (143, 37) \\
& --- I\#2\# Reactor Main Steam System (1214, 302) \\
& --- LBF20AA102 (1199, 599) \\
& --- LBF20AA202 (647, 458) \\
\bottomrule
\end{tabular*}
\end{table}


\section{Discussion} \label{Discussion}

\subsection{Strengths of the Framework}

The NuHF-Claw framework offers several transformative strengths that advance HRA and agent autonomy in digital NPP environments. By moving beyond static, retrospective evaluations, this integrated approach provides a robust methodology specifically tailored to the cognitive challenges of "soft control" behaviors.

\begin{itemize}
    \item \textbf{Continuous Perception and Simulator-Driven Data Acquisition.} A foundational strength of NuHF-Claw lies in its active Perception Layer. Evolving beyond offline data sampling, this framework leverages deep learning (e.g., LSTM networks) to ingest real-time continuous telemetry data from the HTR-PM600 simulator. This enables the instantaneous diagnosis of faults and the identification of initial events (e.g., "Disconnection of Generator to 6kV 1B Bus bar"). This continuous data stream ensures that the subsequent agent layers are always grounded in the precise, real-time physical state of the plant \cite{xiao2024emergency}.

    \item \textbf{Automated Task Modeling via the Procedure-Interface Agent.} The framework resolves the long-standing bottleneck of manual procedure modeling by utilizing the AutoGraph mechanism. It automatically parses natural language emergency operating procedures (EOPs) and maps them onto a dynamic IE-KG. Crucially, this graph embeds detailed spatial coordinate locations and semantic labels of the soft-control interface elements. This transforms static textual procedures into precise, machine-executable operational paths, bridging the semantic gap between written protocols and digital screen navigation.

    \item \textbf{Persistent Cognitive Twin and Dynamic Risk Integration.} The most significant methodological breakthrough resides in the Cognitive Twin and Dynamic Risk Agents. The traditional complexity of manually authoring Lisp code for cognitive modeling is entirely bypassed. Instead, the framework employs large language models as an automated toolchain to dynamically generate ACT-R Lisp code based on real-time operational paths. This creates a continuously running Human Digital Twin capable of accurately simulating operator reaction times—a critical parameter for HRA—and predicting cognitive bottlenecks in real-time. Concurrently, the KRAIL-based module performs a dynamic evaluation of Performance Influencing Factors (PIFs) governed by the IDHEAS macro-cognitive theory, allowing the agent to continuously assess the cognitive failure probability ($P_c$).

    \item \textbf{Risk-Constrained Governance via Bayesian Networks.} Finally, NuHF-Claw does not merely output predictions; it governs actions.  By mapping the multi-step operational procedures into a Bayesian Network, the Governance Layer holistically couples the cognitive failure probability ($P_c$) and the time-dependent failure probability ($P_t$). This provides a real-time, comprehensive measure of the systemic Human Error Probability (HEP). When the dynamic HEP exceeds safety thresholds, the Governance Layer actively intercepts autonomous execution, enforcing a "human-in-the-loop" confirmation to preserve human decision-making authority.
\end{itemize}

\subsection{Comparison with Existing Dynamic HRA Models}

To fully contextualize the theoretical and practical advancements of NuHF-Claw, it is essential to compare it against established dynamic HRA methodologies and previous diagnostic frameworks, particularly HUNTER and DRIF.

\textbf{Comparison with HUNTER:} HUNTER represents a significant milestone in third-generation dynamic HRA, explicitly designed to couple a "virtual operator" with a simulated nuclear system to generate predictive empirical data, fundamentally shifting HRA from a retrospective to a predictive discipline. However, HUNTER faces severe practical limitations regarding scalability and entry barriers \cite{boring2023procedure}. Constructing the virtual operator in HUNTER requires analysts to possess extensive expertise in cognitive modeling and manual coding of procedural execution scripts. Furthermore, it lacks the flexibility to autonomously adapt to novel EOPs or dynamic changes in digital HMIs. 

NuHF-Claw overcomes these limitations through intelligent automation. While HUNTER relies on manually predefined tasks, NuHF-Claw's Procedure-Interface Agent (AutoGraph) automatically parses EOPs and maps them to interface coordinates \cite{xiao2025autograph}. More importantly, NuHF-Claw eliminates the manual coding bottleneck by utilizing LLMs to dynamically generate the underlying ACT-R Lisp code. This transforms the "virtual operator" from a statically coded simulation tool into a persistent, adaptable "always-on" AI entity that continuously updates its cognitive state based on real-time plant telemetry.

\textbf{Comparison with DRIF:}
DRIF is a highly capable framework primarily focused on simulator-driven data acquisition and deep learning-based fault diagnosis \cite{xiao2025dynamic}. It excels at identifying initial events and mapping static, downstream consequences using data-driven algorithms. However, DRIF's architecture is fundamentally a perception and diagnosis tool; it models the \textit{plant's} behavior but treats the \textit{operator} as a largely opaque entity. It lacks a mechanistic understanding of the operator's cognitive trajectory and cannot predict interface-induced errors such as mode confusion or visual search delays.

NuHF-Claw builds upon the perceptual foundation established by diagnostic tools like DRIF but introduces a paradigm shift towards "human-centered autonomy." Where DRIF stops at identifying a fault, NuHF-Claw initiates a cognitive and probabilistic risk assessment loop. By integrating the Cognitive Twin Agent, NuHF-Claw not only knows what is wrong with the plant but also understands how the human operator is perceiving and reacting to the anomaly. Furthermore, unlike traditional diagnostic frameworks, NuHF-Claw acts as an active, risk-constrained co-pilot, utilizing its Governance Layer to suggest interventions or enforce safety gates rather than merely displaying diagnostic results.

\section{Conclusion and Future Research}\label{Conclusion and future research}

The comprehensive transition of nuclear power plant main control rooms (MCRs) from analog panels to highly integrated digital systems has introduced profound operational efficiencies, but it has simultaneously precipitated complex "soft control" behaviors. In these environments, operators engage in intricate, multi-layered information processing, which has inadvertently spawned novel cognitive error modes such as mode confusion, navigational disorientation, and visual search fatigue. Traditional, static Human Reliability Analysis (HRA) methodologies—often constrained by retrospective data and manual modeling bottlenecks—are fundamentally ill-equipped to address the dynamic, real-time nature of these cognitively demanding scenarios. 

To bridge this critical methodological gap, this work introduced \textbf{NuHF-Claw}, a persistent cognitive-risk agent framework engineered specifically for human-centered autonomy in safety-critical nuclear operations. NuHF-Claw transcends conventional diagnostic algorithms by integrating a robust four-layer multi-agent architecture.  Its primary breakthrough lies in overcoming the historical barrier of cognitive modeling: it leverages large language models to automatically generate ACT-R Lisp code, thereby instantiating a continuously running, real-time Human Digital Twin. 

By seamlessly integrating simulator-driven data acquisition (Perception Layer), AutoGraph-based procedural mapping (Procedure-Interface Agent), dynamic execution time simulation (Cognitive Twin Agent), and KRAIL-based real-time PIF analysis (Dynamic Risk Agent), NuHF-Claw achieves a holistic, continuous evaluation of human error probabilities (HEPs). Crucially, through its Bayesian Network-driven Governance Layer, the framework ensures that autonomous AI recommendations are strictly constrained by probabilistic safety gates, effectively mitigating LLM "hallucinations" and firmly preserving the human operator's ultimate decision-making authority.

Building upon this foundational architecture, future research will propel the NuHF-Claw ecosystem in several strategic directions to further solidify its applicability in real-world nuclear operations:

\begin{itemize}
    \item \textbf{Multimodal Cognitive State Calibration:} Future iterations will expand the Perception Layer to incorporate real-time physiological telemetry, such as eye-tracking, pupil dilation, and EEG data. This multimodal integration will provide high-fidelity empirical grounding for the Cognitive Twin Agent, refining its real-time workload and situational awareness predictions beyond simulator interaction logs.
    
    \item \textbf{Algorithmic Latency Optimization:} As the agent runtime heavily relies on LLM orchestration and dynamic ACT-R code generation, minimizing inference latency is paramount for safety-critical control. Future work will focus on deploying localized, lightweight surrogate models and optimizing the runtime environment to ensure microsecond-level responsiveness during rapidly evolving severe accidents.
    
    \item \textbf{Scaling to Multi-Agent Crew Governance:} While the current prototype effectively models individual operator trajectories, future research will expand the framework to capture complex team dynamics. By scaling NuHF-Claw into a multi-operator risk governance system, it will model inter-team coordination, communication failures, and distributed cognitive load, aligning with the complete spectrum of IDHEAS macro-cognitive functions.
    
    \item \textbf{Closed-Loop Simulator Deployment:} We aim to deploy the NuHF-Claw runtime in long-term, closed-loop simulator campaigns with active control room crews. This will generate the extensive empirical validation required to transition the framework from a research prototype to an industrial-grade, risk-informed decision support platform.
\end{itemize}

Ultimately, NuHF-Claw represents a vital paradigm shift in nuclear safety—transitioning from the passive prediction of human errors to the proactive, AI-governed safeguarding of human cognition.

\printcredits

\appendix

\bibliographystyle{cas-model2-names}

\bibliography{nuclearClaw}

\end{document}